\documentclass[lettersize,journal]{IEEEtran}
\usepackage{amsmath,amsfonts}
\usepackage{algorithmic}
\usepackage{algorithm}
\usepackage{array}
\usepackage{textcomp}
\usepackage{stfloats}
\usepackage{url}
\usepackage{hyperref}
\usepackage{verbatim}
\usepackage{amssymb}
\usepackage{graphicx}
\usepackage{cite}
\usepackage{booktabs}       
\usepackage{multirow}       
\usepackage{makecell}       
\usepackage{threeparttable} 
\usepackage{graphicx}
\usepackage{subcaption}

\usepackage{pifont}
\usepackage[dvipsnames]{xcolor}

\usepackage{algorithmic}
\usepackage{algorithm}
\usepackage{array}
\usepackage{verbatim}
\usepackage{threeparttable} 
\usepackage[table]{xcolor}
\usepackage[dvipsnames]{xcolor}

\definecolor{TableBlue}{RGB}{210, 230, 255}

\usepackage[nopatch=footnote]{microtype}
\begin{document}

\title{CMA-OT: Hierarchical Expert Supervision for Dance-to-Music Generation}

\author{

Demo Page: \url{https://beria-moon.github.io/CMA-OT/}
\vspace{1em}

Jinting~Wang$^{1,2}$,
Chenxing~Li$^{2}$,
Dong~Yu$^{2}$,
Li~Liu$^{1,*}$%

\vspace{1em}
$^{1}$HKUST(GZ), $^{2}$Tencent 

\thanks{$^{*}$ Corresponding author: Li LIU, avrillliu@hkust-gz.edu.cn}
}

\markboth{Journal of \LaTeX\ Class Files}%
{Shell \MakeLowercase{\textit{et al.}}: A Sample Article Using IEEEtran.cls for IEEE Journals}


\maketitle

\begin{abstract}
Dance-to-music (D2M) generation aims to synthesize music that is rhythmically and stylistically aligned with dance videos. A key challenge arises from the semantic mismatch between sparse dance cues, such as rhythm and style, and the dense information required for music composition, including structure, instrumentation, and expressive dynamics. Existing methods typically rely on these sparse cues and supervise only the final audio output, resulting in poorly learned music representations and generated music with limited musicality and structural coherence.
To address these issues, we propose
\textbf{C}urriculum-guided \textbf{M}ulti-scale representation \textbf{A}lignment framework with scale-aware \textbf{O}ptimal \textbf{T}ransport (CMA-OT), a novel paradigm that leverages an external music expert to provide hierarchical supervision for the generator’s latent features, bridging the semantic gap and enhancing representation learning.
To effectively incorporate hierarchical supervision, we introduce a curriculum-guided multi-scale learning strategy that progressively transfers musical knowledge from expert to the music generator, enabling stable and effective representation learning. Moreover, to accommodate the semantic and structural variations across different expert scales and achieve fine-grained alignment under temporal mismatch, we propose a scale-aware optimal transport alignment mechanism, which models soft correspondences between hierarchical expert representations and generator's latent features.
Extensive experiments on two datasets demonstrate that CMA-OT achieves state-of-the-art (SOTA) performance in rhythmic synchronization, perceptual quality, and overall music generation.
\end{abstract}





\section{Introduction}
\label{sec:intro}

\begin{figure}[t]
    \centering
    \includegraphics[width=\columnwidth]{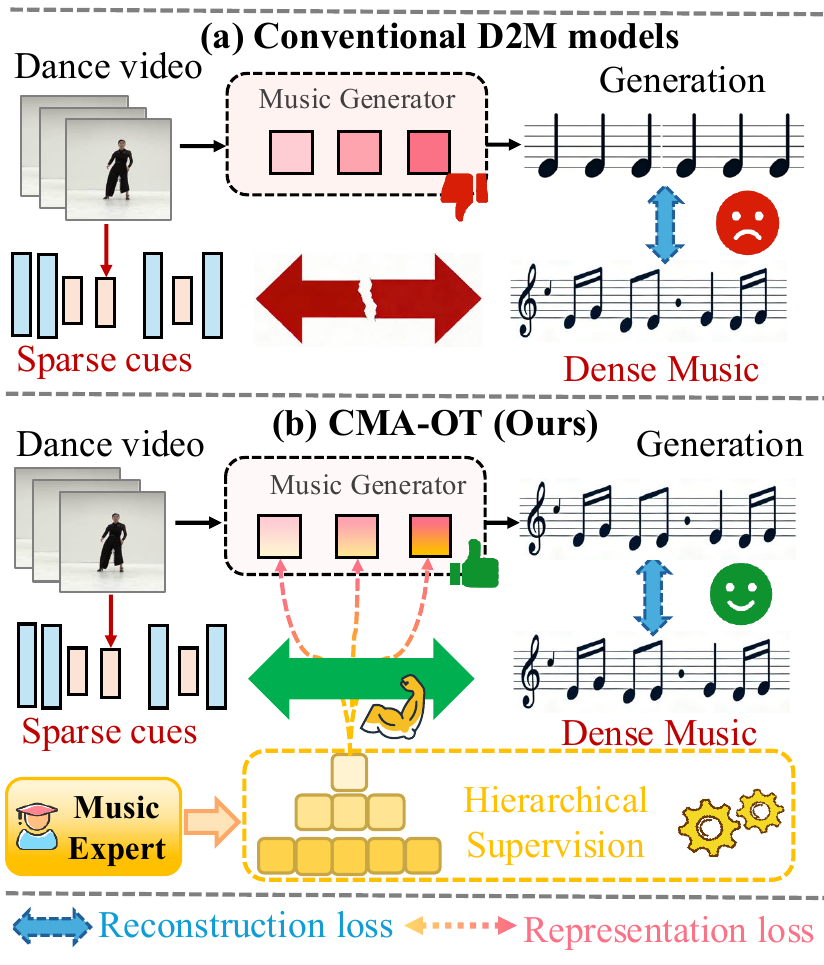}
    \caption{Comparison of D2M paradigms.
(a) Conventional D2M methods rely solely on sparse dance cues and final-output supervision, resulting in inefficient representation learning and poor music generation.
(b) In contrast, CMA-OT leverages hierarchical representation supervision from a pre-trained music expert to bridge the semantic mismatch, enabling coherent, expressive, and well-aligned music generation.}
    \label{fig:motivation}
\end{figure}

Dance-to-music (D2M) generation aims to synthesize music that is rhythmically and stylistically aligned with dance videos, and has attracted increasing attention due to its broad applications in virtual performance and multimedia content creation.

Despite recent progress, D2M remains fundamentally challenging due to the inherent semantic mismatch between dance and music modalities. 
Dance videos provide only sparse cues, such as rhythm and style, while music composition requires dense information, including macro-level structures (\textit{e.g.,} global style), meso-level attributes (\textit{e.g.,} melody), and micro-level details (\textit{e.g.,} rhythm and instrument timbre).
This fundamental gap makes it difficult for models to generate musically coherent, expressive audio from sparse dance inputs alone.
As illustrated in Figure~\ref{fig:motivation}(a), 
most existing D2M methods~\cite{wang2025motioncomposer,sun2025enhancing} 
heavily rely on sparse dance cues and optimize the generator only on the final audio output via reconstruction losses. 
Consequently, these models are forced to simultaneously handle rhythmic alignment and music composition. 
This dual burden, coupled with the scarcity of high-quality dance–music paired data, hinders the learning of meaningful and generalizable music representations. 
As a result, the generated music often lacks musicality, structural coherence, and expressive richness.

Recently, REPA \cite{yurepresentation} has emerged as a method to enhance the semantic representation ability of generative models by aligning representations between diffusion models and pre-trained foundation models \cite{zhang2025videorepa,shan2025hunyuanvideofoley,ton2025taro}. 
This inspires us to introduce external musical knowledge to address the above limitations in D2M.
To this end, we propose a novel learning paradigm: \textbf{hierarchical expert supervision}.
Moving beyond sparse dance cues and final-output-only supervision, we leverage a pre-trained multi-scale music expert to provide hierarchical supervision on the latent representation space of the music generator.
As illustrated in Figure~\ref{fig:motivation}(b), the expert acts like a \textit{composer consultant} that offers coarse-to-fine supervision from high-level structures to fine-grained temporal dynamics, while the dance video serves as a \textit{conductor} to ensure rhythmic and stylistic consistency.
This paradigm effectively bridges the semantic mismatch between dance and music and enables robust music representation learning, leading to coherent, expressive, and well-aligned music generation.
However, directly adopting multi-scale expert supervision faces new challenges.
\textbf{First}, multi-scale expert signals vary significantly in semantics and structure, and simultaneously incorporating multiple expert scales may induce training instability and optimization imbalance.
\textbf{Second}, fine-grained representation alignment requires scale-aware structure modeling rather than global semantic matching, as the expert representations and generator latent features often have mismatched temporal resolutions across hierarchical scales.

To tackle these challenges, we propose \textbf{CMA-OT}, a \textbf{C}urriculum-guided \textbf{M}ulti-scale representation 
\textbf{A}lignment framework with scale-aware \textbf{O}ptimal 
\textbf{T}ransport. 
\textbf{First}, to stabilize knowledge transfer across hierarchical levels, we design a curriculum-guided multi-scale learning strategy that progressively injects supervision from coarse to fine, avoiding conflicts between scales and ensuring stable representation learning.
\textbf{Second}, to adapt to scale-varying semantics and enable fine-grained soft alignment, we propose a scale-aware Fused Gromov-Wasserstein (FGW) alignment mechanism. It models flexible correspondences between expert representations and generator latents while preserving both semantic consistency and temporal structure, outperforming conventional global alignment methods.

Our main contributions are summarized as follows:

\begin{itemize}
    \item To address the challenge of semantic mismatch between sparse dance cues and dense musical semantics, we propose CMA-OT, a hierarchical expert supervision paradigm that leverages a pre-trained multi-scale music expert, providing structured guidance across multiple semantic levels.
    \item We introduce two key components: a curriculum-guided multi-scale learning strategy for stable, progressive knowledge transfer, and a scale-aware FGW alignment mechanism for flexible, fine-grained cross-scale alignment.
    \item Extensive experiments show that CMA-OT outperforms state-of-the-art (SOTA) approaches on two datasets, validating the effectiveness of hierarchical expert supervision for high-quality D2M generation.
\end{itemize}

\section{Related Work}
\label{sec:related_work}

\subsection{Dance-to-Music Generation}

D2M generation aims to synthesize music conditioned on dance inputs, with existing methods primarily relying on dance-centric representations. 
These approaches ~\cite{zhu2022quantized,sun2025enhancing,liang2024dancecomposer} typically extract motion features using pre-trained encoders, \textit{i.e.,} STGCN \cite{yan2018spatial} or leverage pose-based rhythmic cues~\cite{su2021does,Yu2023Long,wang2025motioncomposer}, followed by optimizing a music generator via reconstruction objectives. 
While effective in capturing global motion semantics and basic rhythmic alignment, such designs inherently depend on sparse dance cues and lack explicit modeling of rich musical structures. 
As a result, the generated music often exhibits limited diversity and weak structural coherence, particularly in terms of harmony, texture, and long-range organization.
These limitations highlight a fundamental gap in existing D2M systems, the absence of a structured supervision paradigm that can inject rich musical knowledge into the generation process. 

In contrast, our work introduces hierarchical representation supervision, where a pre-trained multi-scale music expert provides structured guidance across semantic levels, enabling both global coherence and fine-grained temporal alignment.

\begin{figure*}
    \centering
    \includegraphics[width=0.9\linewidth]{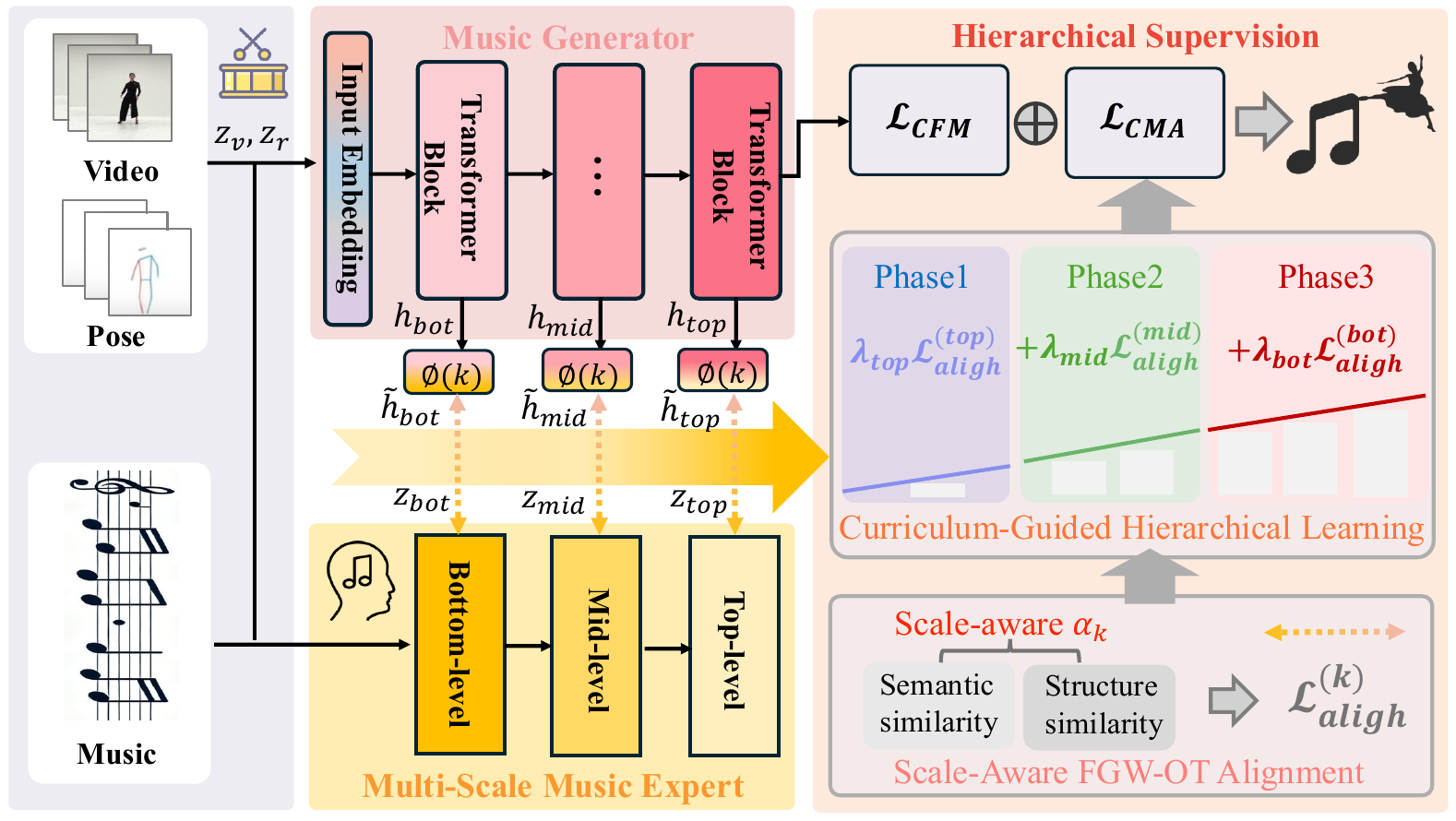}
    \caption{\textbf{Overview of CMA-OT.}
It is a novel paradigm-driven hierarchical supervision framework for D2M.
Given an input dance video, the framework extracts visual style and temporal rhythm features to condition a Diffusion Transformer.
A curriculum-guided multi-scale strategy is adopted to progressively distill hierarchical knowledge from a pre-trained music expert, and scale-aware FGW-OT is further introduced to enforce temporally adaptive representation alignment.
Benefiting from this hierarchical supervision paradigm, our model achieves accurate dance-music synchronization and high-fidelity musical generation.
}
    \label{fig:pipeline}
\end{figure*}

\subsection{Representation Alignment for Generative Models}

Recent works have explored representation alignment to enhance generative models by transferring knowledge from pre-trained encoders. 
REPA~\cite{yurepresentation} aligns internal representations between diffusion models and pre-trained vision encoders, improving both semantic consistency and training efficiency. 
Subsequent works extend this idea to multimodal settings. For instance, JanusFlow~\cite{ma2025janusflow} aligns intermediate features between generation and understanding modules, while TARO~\cite{ton2025taro} and HunyuanVideo-Foley~\cite{shan2025hunyuanvideofoley} apply representation alignment to sequential audio generation tasks. 
VideoREPA~\cite{zhang2025videorepa} further distills physics-aware knowledge from video understanding foundation models into text-to-video (T2V) generators.
However, existing alignment methods are primarily designed to enforce global semantic consistency or point-wise feature matching. 
They typically rely on direct similarity objectives (\textit{e.g.}, cosine similarity) and do not explicitly model temporal structures or hierarchical relationships. 
Such limitations make them less suitable for tasks such as D2M generation, where fine-grained temporal dynamics and multi-scale dependencies are essential. 

Conversely, our work departs from global alignment and introduces a hierarchical, scale-aware alignment paradigm that enables structured knowledge transfer from multi-scale music experts, leading to improved temporal coherence and musical expressiveness.

\subsection{Optimal Transport for Cross-Space Alignment}

Optimal Transport (OT)~\cite{peyre2019computational} provides a principled framework for aligning probability distributions. 
The Wasserstein distance~\cite{thorpe2018introduction} measures discrepancies based on feature similarity in a shared space, while Gromov-Wasserstein distance~\cite{memoli2011gromov} extends this formulation to compare relational structures across heterogeneous domains. 
FGW~\cite{vayer2019optimal} further integrates both feature similarity and structural consistency, enabling alignment that preserves semantic and relational information simultaneously.
FGW distance has been successfully applied in various tasks, including graph matching~\cite{ma2023fused,qian2024reimagining}, action segmentation~\cite{xu2024temporally,luo2022weakly}, and cross-modal alignment~\cite{rahman2025spase,khan2025comprehensive}.

However, its application in generative modeling, particularly for hierarchical and scale-aware supervision, remains underexplored.
In this work, we extend FGW distance into a scale-aware alignment mechanism, enabling flexible correspondence modeling across multiple semantic levels.


\subsection{Curriculum Learning for Structured Training}

Curriculum learning~\cite{bengio2009curriculum} improves training stability and generalization by organizing learning in a progressive manner, where models are first exposed to simpler patterns and gradually introduced to more complex structures. This learning paradigm has been widely adopted in generative modeling~\cite{na2025boost,frolov2024objblur,huang2020dance}, reinforcement learning~\cite{croitoru2025curriculum,uppuluri2025curla,wen2025sari}, and segmentation tasks~\cite{wang2023grenet,dai2024curriculum,liu2025sscl}. These approaches demonstrate that curriculum-based optimization can effectively mitigate optimization difficulty, reduce training instability, and improve model generalization.

Despite its success, prior works typically apply curriculum learning at the data or task level, without explicitly considering hierarchical representation learning or multi-scale supervision.
In this work, we introduce a structured curriculum learning strategy that operates at the representation level. Specifically, our method progressively transfers supervision from coarse semantic structure to fine-grained temporal dynamics, enabling the model to learn global musical organization before refining local details. This hierarchical learning process reduces optimization complexity and encourages more coherent long-term structure generation.

\section{Preliminary}
\label{sec:preliminary}

This section establishes the theoretical foundations of our framework, focusing on hierarchical music representations and the FGW distance for structured representation alignment.

\subsection{Multi-Scale Music Expert}
To bridge the semantic gap between sparse dance cues and rich musical structures, we leverage a pre-trained multi-scale VQ-VAE encoder (\textit{e.g.,} from Jukebox~\cite{dhariwal2020jukebox}). This expert model captures hierarchical latent representations at multiple temporal resolutions via a cascaded architecture. 

Formally, the encoder maps raw audio into three discrete latent levels: top-level codes $\mathbf{z}_{\text{top}}$ that encode global musical semantics such as style and genre; middle-level codes $\mathbf{z}_{\text{mid}}$ that capture intermediate structures including melodic and harmonic phrases; and bottom-level codes $\mathbf{z}_{\text{bot}}$ that represent fine-grained rhythmic textures and timbre. We denote the set of hierarchical expert priors as $\mathcal{Z}_{\text{exp}} = \{\mathbf{z}_k\}_{k \in \{\text{top, mid, bot}\}}$, providing comprehensive guidance from global composition to local temporal dynamics.

\subsection{Fused Gromov-Wasserstein Distance}
Optimal Transport (OT)~\cite{monge1781memoire} offers a principled approach for aligning distributions, inherently supporting sequences of unequal lengths, which is essential for aligning hierarchical music expert features across varying temporal resolutions.
Standard OT primarily considers global feature similarity and ignores relational structures within sequences. FGW~\cite{vayer2019optimal} addresses this limitation by jointly optimizing feature correspondence and structural consistency. Compared to conventional similarity metrics such as cosine similarity, which focus solely on global feature correspondence, FGW distance additionally captures relational structures within sequences. This makes FGW distance particularly suitable for representation alignment tasks where structural consistency is important.

Given a source sequence $\mathbf{X} = \{\mathbf{x}_i\}_{i=1}^n$, a target sequence $\mathbf{Y} = \{\mathbf{y}_k\}_{k=1}^m$, and their intra-sequence relation matrices $\mathbf{C}^X \in \mathbb{R}^{n \times n}$, $\mathbf{C}^Y \in \mathbb{R}^{m \times m}$, the FGW distance is defined as:

\begin{equation}
\begin{aligned}
\mathcal{D}_{\mathrm{FGW}}(\mathbf{X}, \mathbf{Y})
= \min_{\mathbf{P} \in \Pi(\mathbf{a}, \mathbf{b})}
\sum_{i,k}
\Big[
(1 - \alpha)\, c(\mathbf{x}_i, \mathbf{y}_k)
\\
+ \alpha
\sum_{j,l}
\mathbf{P}_{j,l}
L(C^{X}_{i,j}, C^{Y}_{k,l})
\Big]
\mathbf{P}_{i,k},
\end{aligned}
\label{eq:standard-fgw}
\end{equation}
where $\mathbf{P} \in \mathbb{R}^{n \times m}$ is the transport plan, $\Pi(\mathbf{a},\mathbf{b})$ denotes the set of valid couplings with marginals $\mathbf{a}$ and $\mathbf{b}$, $c(\cdot)$ measures feature-level discrepancy, $L(\cdot)$ penalizes distortions of pairwise relations, and $\alpha \in [0,1]$ balances feature alignment and structural preservation. 

\section{Method}
\label{sec:method}

\subsection{Overview}
To generate music that is both rhythmically aligned and musically coherent with dance videos, we propose \textbf{CMA-OT}, a paradigm-driven framework that introduces hierarchical expert supervision. Beyond simply utilizing a pre-trained multi-scale music expert, CMA-OT designs a dedicated alignment pipeline to fully exploit its multi-scale structural knowledge.

As illustrated in Figure.~\ref{fig:pipeline}, given a dance video and its pose sequences, an I3D visual encoder~\cite{carreira2017quo} extracts global style features $\mathbf{z}_v \in \mathbb{R}^{T_v \times D_v}$, while a rhythm encoder~\cite{wang2025gaca} captures fine-grained temporal dynamics to produce $\mathbf{z}_r \in \mathbb{R}^{T_r \times D_r}$, where $T_v$ and $T_r$ denote the temporal dimensions, and $D_v, D_r$ are feature dimensions. These dance-derived features condition a flow-based music generator acting as a temporal conductor to ensure rhythmic alignment, with the pre-trained music expert providing hierarchical structural guidance for musical coherence.

\subsection{Curriculum-Guided Multi-Scale Alignment}
\label{subsec:curriculum_alignment}

To bridge the hierarchical features of the pre-trained music expert with the DiT backbone, we propose a curriculum-guided multi-scale alignment scheme. This approach enforces semantic and structural consistency in a coarse-to-fine manner, ensuring progressively transfer expert knowledge to music generation.

\noindent \textbf{Multi-scale Feature Mapping.} The music expert provides hierarchical representations capturing musical cues from global structures to local details. We establish coarse-to-fine alignment across three semantic levels:
\begin{itemize}[leftmargin=*, noitemsep]
    \item \textbf{Coarse-level:} Global style features $\mathbf{z}_{\text{top}}$ are aligned with deep-layer DiT representations $\mathbf{h}_{\text{top}}$ to anchor high-level musical semantics.
    \item \textbf{Medium-level:} Structural and melodic features $\mathbf{z}_{\text{mid}}$ are aligned with middle-layer DiT representations $\mathbf{h}_{\text{mid}}$ to enhance compositional coherence.
    \item \textbf{Fine-level:} Local rhythmic patterns $\mathbf{z}_{\text{bot}}$ are aligned with shallow-layer DiT representations $\mathbf{h}_{\text{bot}}$ to refine fine-grained temporal dynamics.
\end{itemize}

Since the expert operates in the clean audio domain while the DiT processes noised latents, we follow REPA~\cite{yurepresentation} and introduce lightweight projection heads $\phi_{(k)}$ (two-layer MLPs with GELU activations) to map internal representations into a comparable semantic space:
\begin{equation}
\tilde{\mathbf{h}}_k = \phi_{(k)}(\mathbf{h}_k), \quad k \in \{\text{top}, \text{mid}, \text{bot}\}.
\end{equation}
The projected representations $\tilde{\mathbf{h}}_k$ are then used to compute the representation alignment loss:
\begin{equation}
\label{eq:alignment_loss}
\mathcal{L}_{\text{align}}^{(k)} = \mathrm{Align}\big(\tilde{\mathbf{h}}_k, \mathbf{z}_k\big),
\end{equation}
where $\mathrm{Align}(\cdot, \cdot)$ denotes a scale-aware distance accounting for both temporal structure and semantic similarity (detailed in Sec.~\ref{subsec:scale_aware_fgw}).

\begin{figure}[t]
    \centering
    \includegraphics[width=1\linewidth]{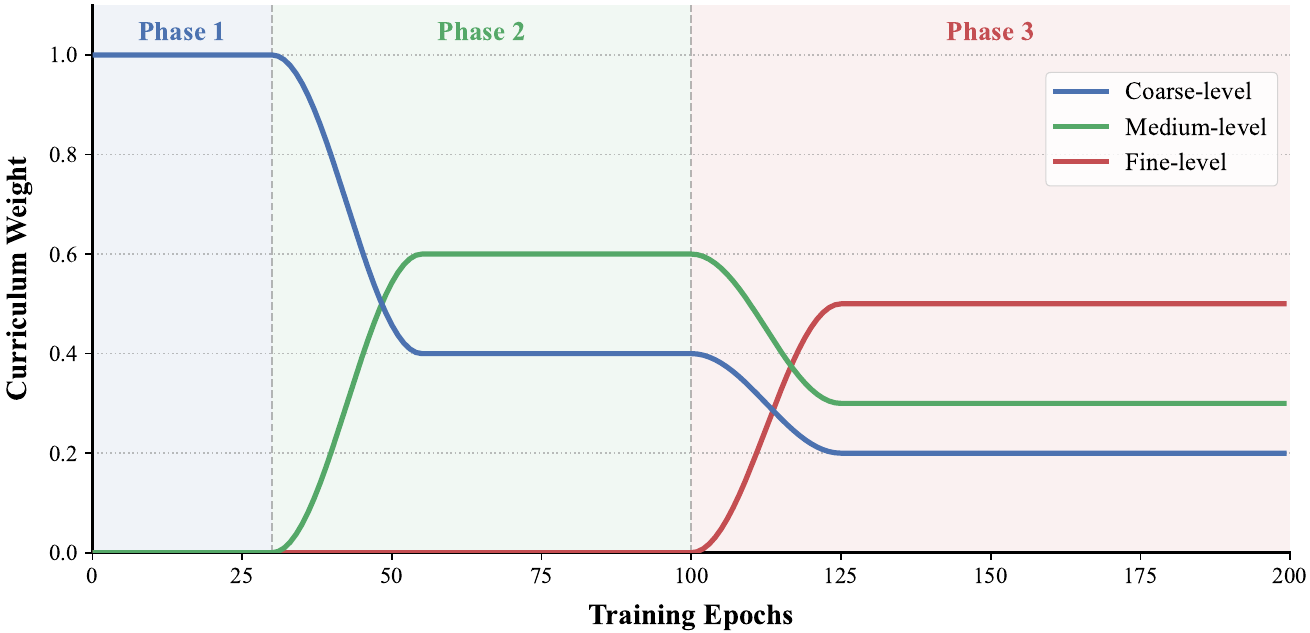} %
    \caption{\textbf{Curriculum Weights Visualization.} Three-phase alignment weights following a cosine annealing schedule, showing the progressive activation of top, middle, and bottom-level representation alignment across training epochs.}
    \label{fig:curriculum_weights}
\end{figure}

\noindent \textbf{Progressive Curriculum Strategy.}
To mitigate conflicts when enforcing alignment across all scales simultaneously, we adopt a smooth three-phase curriculum strategy with cosine-annealed curriculum weight $\lambda_k$, as illustrated in Figure~\ref{fig:curriculum_weights}.
The curriculum progressively activates the multi-scale alignment loss $\mathcal{L}_{\text{align}}^{(k)}$ in a coarse-to-fine order:
\begin{itemize}[leftmargin=*, noitemsep]
    \item \textbf{Phase 1:}  The model focuses on top-level expert alignment and optimizes
    $\lambda_{top }*\mathcal{L}_{\text{align}}^{(\text{top})}$ to capture global musical style and high-level semantics.
    \item \textbf{Phase 2:} Middle-level expert representation alignment is gradually introduced while the top-level weight is softly reduced. The model optimizes $\lambda_{top }*\mathcal{L}_{\text{align}}^{(\text{top})}+ \lambda_{mid }*\mathcal{L}_{\text{align}}^{(\text{mid})}$ to learn melodic and harmonic structures. 
    \item \textbf{Phase 3:} Bottom-level expert representation alignment becomes dominant to refine fine-grained temporal and rhythmic precision, while higher-level weights decay smoothly. The model optimizes the full objective $\lambda_{top }*\mathcal{L}_{\text{align}}^{(\text{top})}+ \lambda_{mid}*\mathcal{L}_{\text{align}}^{(\text{mid})}+\lambda_{bot}*\mathcal{L}_{\text{align}}^{(\text{bot})}$,
    to maintain optimization stability.
\end{itemize}

This progressive shift effectively alleviates gradient interference and ensures consistent knowledge transfer across hierarchical scales.
The total curriculum-guided multi-scale alignment loss $\mathcal{L}_{\text{CMA}}$ is computed at the final denoising step, when DiT representations approximate the clean data most closely:
\begin{equation}
\mathcal{L}_{\text{CMA}} = \sum_{k \in \{\text{top}, \text{mid}, \text{bot}\}} \lambda_k \, \mathcal{L}_{\text{align}}^{(k)} \Big.
\end{equation}
This formulation ensures supervision is applied in a perceptually meaningful space, directly comparable to clean expert features.

\begin{table*}[t]
\centering
\caption{Comprehensive quantitative comparison on AIST++ and TikTok datasets. The best results are highlighted in bold, while second-best are underlined.}
\label{tab:comparison_all}
\begin{tabular}{l|ccccc|cccc|cc}
\toprule
\multirow{2}{*}{Method} 
& \multicolumn{5}{c|}{Rhythm Alignment} 
& \multicolumn{4}{c|}{Aesthetic Quality} 
& \multicolumn{2}{c}{Music Quality} \\
\cmidrule(lr){2-6} \cmidrule(lr){7-10} \cmidrule(lr){11-12}
 & BCS$\uparrow$ & CSD$\downarrow$ & BHS$\uparrow$ & HSD$\downarrow$ & F1$\uparrow$
 & CE$\uparrow$ & CU$\uparrow$ & PC$\uparrow$ & PQ$\uparrow$
 & FAD$_p\downarrow$ & FAD$_c\downarrow$ \\
\midrule
\multicolumn{12}{c}{\textbf{AIST++}} \\
\midrule
D2M-GAN \cite{zhu2022quantized}  & 89.09 & 14.05 & 88.95 & 22.23 & 88.84 & 4.76 & 6.65 & 5.13 & 6.33 & 49.49 & 1.27  \\
CDCD \cite{zhu2022discrete}      & 92.18 & 14.66 & 89.50 & 21.16 & 88.95 & 6.69 & 6.76 & 5.31 & 6.30 & 47.95 & 1.25  \\
LORIS \cite{Yu2023Long}      & 92.13 & 7.94  & 91.71 & \underline{9.69}  & 92.05 & 6.41 & 6.80  & 6.23 & 6.74  & 33.64 & 0.68  \\
Textual-Inv \cite{li2024dance}&93.21& 11.74& 84.97& 17.15&86.75&6.73& 7.26& \underline{6.56}& 7.10& 28.70&0.52\\
MotionComposer \cite{wang2025motioncomposer} & \underline{95.84} & \underline{7.89}  & \underline{95.09} & 16.09 & \underline{96.45} & \underline{6.87} & \underline{7.46}  & 6.49 & \underline{7.33} & \underline{27.52} & \underline{0.49}  \\
CMA-OT (Ours)                     & \textbf{99.14} & \textbf{7.07} & \textbf{99.12} & \textbf{7.56} & \textbf{99.12} & \textbf{6.93} & \textbf{7.49} & \textbf{6.71} & \textbf{7.44} & \textbf{12.50} & \textbf{0.26} \\
\midrule
\multicolumn{12}{c}{\textbf{TikTok}} \\
\midrule
D2M-GAN \cite{zhu2022quantized}  & 82.55 & 28.03 & 87.07 & 24.62 & 84.22 & 4.49 & 5.61 & 5.24 & 5.84 & 52.68 & 1.48 \\
CDCD \cite{zhu2022discrete}      & 84.46 & 25.95 & 87.95 & 23.27 & 86.45 & 5.12 & 5.91 & 5.32 & 5.94 & 49.55 & 1.36 \\
LORIS \cite{Yu2023Long}      & 88.02 & 24.14 & 85.25 & 18.72 & 87.43 & 5.64 & 6.81 & 6.13 & 6.52 & 45.18 & 0.94 \\
Textual-Inv \cite{li2024dance}&\underline{90.16}&27.41&82.08& 21.62&  85.43&\underline{6.21}&7.10& 6.06& \underline{7.28}& \underline{29.64}& \underline{0.78}\\
MotionComposer \cite{wang2025motioncomposer} & 89.09 & \underline{17.95} & \underline{90.02} & \underline{17.35} & \underline{89.50} & 6.19 & \underline{7.15} & \underline{6.41} &{7.24} & 30.61 & 0.81  \\
CMA-OT (Ours)                     & \textbf{92.88} & \textbf{13.42} & \textbf{92.05} & \textbf{16.27} & \textbf{92.41} & \textbf{6.31} & \textbf{7.19} & \textbf{6.67} & \textbf{7.36} & \textbf{27.53} & \textbf{0.38} \\
\bottomrule
\end{tabular}%

\end{table*}

\subsection{Scale-Aware Optimal Transport Alignment}
\label{subsec:scale_aware_fgw}

Although the curriculum-guided multi-scale alignment provides effective supervision for hierarchical expert representations, semantic and temporal discrepancies across different scales of the external music prior, as well as fine-grained misalignment with the DiT internal states, remain challenging. To address this, we propose a scale-aware representation alignment mechanism. In this scheme, the FGW distance jointly models semantic correspondence and structural relations to resolve temporal misalignment, while a scale-aware strategy adaptively balances these constraints across levels, thereby enhancing rhythmic consistency and semantic coherence. 

Formally, we formulate the alignment loss in Eq.~\eqref{eq:alignment_loss} as a scale-aware FGW objective:
\begin{equation}
\begin{aligned}
\mathcal{L}_{\text{align}}^{(k)}
= \min_{\mathbf{P}^{(k)} \in \Pi(\mathbf{a},\mathbf{b})}
\sum_{i,p} \Bigg[ 
(1 - \alpha_k)\, c\big(\tilde{\mathbf{h}}_{k,i}, \mathbf{z}_{k,p}\big) 
\\[2pt]
+ \alpha_k \sum_{j,q} \mathbf{P}_{j,q}^{(k)} 
\, L\big(\mathbf{C}_{i,j}^{\tilde{\mathbf{h}}_k}, \mathbf{C}_{p,q}^{\mathbf{z}_k}\big)
\Bigg] 
\mathbf{P}_{i,p}^{(k)}
\end{aligned}
\label{eq:scale-aware-fgw}
\end{equation}
where $\mathbf{P}^{(k)}$ denotes the transport plan at the $k$-th scale,
$\mathbf{C}^{\tilde{\mathbf{h}}_k}$ and $\mathbf{C}^{\mathbf{z}_k}$ denote the intra-sequence relation matrices
that preserve the internal topological structure of the two sequences.

To adaptively balance semantic and structural preservation, we define $m_k$ and $n_k$ as the expected semantic and structural costs under the current transport plan $\mathbf{P}^{(k)}$:
\begin{equation}
m_k \triangleq \mathbb{E}_{\mathbf{P}^{(k)}}
\left[c(\tilde{\mathbf{h}}_k, \mathbf{z}_k)\right],
\quad
n_k \triangleq \mathbb{E}_{\mathbf{P}^{(k)}}
\left[L(\mathbf{C}^{\tilde{\mathbf{h}}_k}, \mathbf{C}^{\mathbf{z}_k})\right].
\end{equation}
The adaptive factor $\alpha_k$ is then computed as:
\begin{equation}
\alpha_k = \sigma \Big( \log \frac{n_k}{m_k + \varepsilon} \Big),
\end{equation}
where $\sigma(\cdot)$ is the sigmoid function and $\varepsilon$ a small constant for numerical stability. 
This formulation preserves the soft-alignment nature of FGW while dynamically weighting semantic and structural consistency at each scale, ensuring high-fidelity music synthesis.

\subsection{Conditional Music Generation}
Our music generator is built upon conditional flow matching~\cite{lipman2023flow}, a stable and high-fidelity generative model that learns a continuous trajectory from Gaussian noise to music latents.  

Let $Z_0$ denote the latent representation of the input music signal obtained via a VAE encoder, and let $Z_t$ represent its noisy version at time step $t$.  
Given dance-derived visual style features $z_v$ and temporal rhythm features $z_r$, the model learns a conditional velocity field:
\begin{equation}
\frac{dZ_t}{dt} = \mathbf{v}_\theta(Z_t, t \mid z_v, z_r), \quad Z_0 \sim p_0(Z), \, Z_1 \sim p_1(Z),
\end{equation}
where $p_0$ is the prior Gaussian distribution and $p_1$ corresponds to the distribution of music latents.
The conditional flow matching loss is defined as:
\begin{equation}
\mathcal{L}_{\text{CFM}} = \mathbb{E}_{t,Z_0,Z_1} \big\| \mathbf{v}_\theta(Z_t, t \mid z_v, z_r) - (Z_1 - Z_0) \big\|^2,
\end{equation}
which trains the velocity field to map noisy latent trajectories back to the target music latent.
The final training objective jointly optimizes the flow-matching generation loss and the hierarchical expert alignment loss:
\begin{equation}
\mathcal{L}_{\text{total}} = \mathcal{L}_{\text{CFM}} + \mathcal{L}_{\text{CMA}}.
\end{equation}
This combined loss ensures that the generated music is both perceptually realistic and rhythmically synchronized with the input dance, while preserving strong hierarchical semantic consistency.

\section{Experiment}
\label{sec:experiment}

\subsection{Dataset}
We use two widly used datasets to evaluate performance: AIST++ \cite{li2021ai} and TikTok \cite{zhu2022quantized}. The AIST++ dataset \cite{li2021ai} comprises 1,020 dance videos spanning 10 dance genres, each paired with corresponding music. All videos are recorded in professional studios with clean backgrounds. The TikTok dataset \cite{zhu2022quantized}, collected from the short-video platform TikTok, contains 445 dance videos covering 85 different songs. During training and evaluation, we adhered to the standard dataset splits and evaluation protocols established in previous works \cite{zhu2022discrete, sun2025enhancing} to ensure fair comparison.

\subsection{Implementation Details}
We use a pre-trained I3D model~\cite{carreira2017quo} as the semantic encoder to extract video features from frames, and a motion encoder to capture rhythmic cues from 2D dance poses. 
Each training sample contains a 5-second music clip sampled at 44.1 kHz, and the corresponding 2D pose skeletons are extracted using DWpose~\cite{yang2023effective} with default hyperparameters.
Music waveforms are encoded and decoded via a pre-trained VAE from DiffRhythm~\cite{ning2025diffrhythm}, which employs 5× downsampling blocks (compression factor $f=2048$) to produce 64-dimensional latent representations at 21.5 Hz.
The conditional DiT backbone comprises 12 LLaMA decoder layers\footnote{https://github.com/huggingface/transformers} with 764 hidden dimensions and 12 self-attention heads, implemented using the Hugging Face Transformers library for consistent and reproducible model deployment, following the implementation paradigm of DiffRhythm~\cite{ning2025diffrhythm}.
For scale-aware FGW alignment, we use cosine distance for the feature cost $c(\cdot)$ and MSE loss for the structural cost $L(\cdot)$. The adaptive balancing factor $\alpha_k$ is computed dynamically per scale during training, with a small constant $\epsilon=10^{-6}$ added to the denominator for numerical stability.

We train the model for 200 epochs with a batch size of 4 using the Adam optimizer ($\beta_1=0.9$, $\beta_2=0.95$) and a learning rate of $1\times 10^{-4}$. 
The curriculum schedule follows a three-stage cosine annealing strategy: the top-level alignment dominates during the first 30 epochs, the mid-level alignment weight gradually increases from epoch 30 to 100, and the bottom-level alignment is activated after epoch 100. Each transition phase lasts for 25 epochs to ensure smooth weight adaptation and avoid training instability.
During inference, the diffusion process is solved using a 32-step Euler ODE sampler with classifier-free guidance~\cite{ho2021classifier} at a scale factor of 4.
Additional implementation details are provided in the supplementary material.



\subsection{Evaluation Metrics}

\textbf{Rhythm Alignment.}
We follow prior work \cite{Yu2023Long} to evaluate the musical alignment. Specifically, we use beats coverage scores (BCS) and beats hit scores (BHS), reporting both their F1 scores and standard deviations (denoted as CSD and HSD). BCS measures the fraction of aligned rhythm points relative to the total beats in the generated music, while BHS measures the proportion of aligned beats with respect to the ground truth.

\noindent\textbf{Music Quality.} 
We evaluate the similarity between generated music and ground truth using the Fr\'echet Audio Distance (FAD)~\cite{roblek2019fr}. In particular, we employ three audio feature extractors: 
(i) PANNs~\cite{kong2020panns}, an audio neural network trained on the large-scale AudioSet dataset~\cite{gemmeke2017audioset}; 
(ii) CLAP~\cite{zhao2023clap}, a contrastive language-audio model trained on the large-scale LAION-Audio-630K dataset~\cite{10095969}.
The corresponding FAD scores are denoted as FAD$_p$, FAD$_c$, respectively.

\noindent\textbf{Aesthetic Quality.} 
To assess the aesthetic quality of generated music, we adopt Meta Audiobox-aesthetics \cite{tjandra2025meta} in four dimensions: Production Quality (PQ), covering clarity, fidelity, dynamics, and spatialization of the audio; Production Complexity (PC), reflecting the richness of audio components; Content Enjoyment (CE), capturing subjective appeal and artistic value; and Content Usefulness (CU), indicating the potential for reuse in content creation.

\noindent\textbf{Subjective Evaluation.}
To enable a comprehensive comparison with competitive approaches, we conducted a user study to evaluate both the perceptual quality and the relevance of the generated music to the dance videos. Volunteers were asked to complete a questionnaire based on music samples generated from the AIST++ dataset. Human raters assessed each sample on two aspects: (i) overall perceptual quality (OVL) and (ii) relevance to the input dance video (REL), using a 5-point Likert scale ranging from 1 (poor) to 5 (excellent).

\begin{table*}[t]
\centering
\caption{
Ablation study on key components of CMA-OT on the AIST++ dataset. 
The best results are highlighted in bold, while
second-best are underlined.
}
\label{tab:ablation}

\begin{tabular}{l|ccccc|cccc|cc}
\toprule
\multirow{2}{*}{Method} 
& \multicolumn{5}{c|}{Rhythm Alignment} 
& \multicolumn{4}{c|}{Aesthetic Quality} 
& \multicolumn{2}{c}{Music Quality} \\
\cmidrule(lr){2-6} \cmidrule(lr){7-10} \cmidrule(lr){11-12}
 & BCS$\uparrow$ & CSD$\downarrow$ & BHS$\uparrow$ & HSD$\downarrow$ & F1$\uparrow$
 & CE$\uparrow$ & CU$\uparrow$ & PC$\uparrow$ & PQ$\uparrow$
 & FAD$_p$$\downarrow$ & FAD$_c$$\downarrow$ \\
\midrule
Base & 97.19 & 9.54 & 97.42 & 9.31 & 97.27 & 6.04 & 6.71 & 6.54 & 6.92 & 21.12 & 0.42 \\
Base + CLAP-OT \cite{elizalde2023clap} & 97.94 & 8.76 & 97.84 & 11.09 & 97.90 & 6.14 & 6.75 & 6.51 & 6.83 & 20.42 & 0.39 \\
Base + MERT-OT \cite{li2024mert}&98.12&8.17&98.23&12.29&98.18&6.33&7.12&6.62&7.13&19.76&0.37  \\
Base + Top-OT & 97.93 & 8.19 & 97.76 & 9.22 & 97.83 & 6.32 & 6.78 & 6.57 & 7.10 & 19.91 & 0.38 \\
Base + Middle-OT & 98.26 & 8.03 & 98.47 & 9.06 & 98.39 & 6.46 & 7.18 & 6.65 & 7.15 & 18.92 & 0.35 \\
Base + Bottom-OT & 98.65 & 7.81 & 98.72 & 8.89 & 98.67& 6.61 & 7.24 & 6.67 & 7.21 & 15.69 & 0.32 \\
\midrule
Base + MA-OT & 97.77 & 8.09 & 97.70 & 12.05 & 97.73 & 6.05 & 6.68 & 6.57 & 6.84 & 20.64 & 0.39 \\
Base + CMA & 98.68 & 7.83 & 98.79 & 8.72 & 98.71 & 6.73 & 7.26 & 6.69 & 7.24 & 13.36 & 0.29 \\
\midrule
Base + CMA-OT* & \underline{99.03} & 7.31 & \underline{99.04} & \underline{7.92} & \underline{99.03} & \underline{6.81} & \underline{7.33} & \underline{6.69} & \underline{7.42} & \underline{12.84} & \underline{0.27} \\
\textbf{Base + CMA-OT (Ours)} & \textbf{99.14} & \textbf{7.07} & \textbf{99.12} & \textbf{7.56} & \textbf{99.12} & \textbf{6.93} & \textbf{7.49} & \textbf{6.71} & \textbf{7.44} & \textbf{12.50} & \textbf{0.26} \\
\bottomrule
\end{tabular}%

\end{table*}

\subsection{Comparison with SOTA Methods}
To validate the effectiveness of our proposed CMA-OT, we compare it with several D2M methods, including D2M-GAN \cite{zhu2022quantized}, CDCD \cite{zhu2022discrete}, LORIS \cite{Yu2023Long}, Textual-Inv \cite{li2024dance}, and MotionComposer \cite{wang2025motioncomposer}.

\noindent\textbf{Quantitative Comparison.} As shown in Table~\ref{tab:comparison_all}, CMA-OT consistently outperforms all baselines across rhythmic alignment, aesthetic quality, and music quality. 
On AIST++, CMA-OT surpasses MotionComposer b in BHS, and in F1 score, demonstrating superior temporal synchronization and rhythmic modeling capabilities. 
Moreover, CMA-OT achieves the best aesthetic scores and significantly reduces FAD$_p$ and FAD$_c$, indicating substantially improved perceptual and structural music quality.
On the more challenging TikTok dataset, CMA-OT maintains consistent superiority, achieving the best performance in rhythmic alignment and competitive aesthetic quality while producing the lowest FAD scores. 
These results demonstrate that our CMA-OT effectively bridges the semantic gap between dance motion and musical semantics, leading to high-quality music generation.

\begin{figure}[ht]
    \centering
    \includegraphics[width=0.99\linewidth]{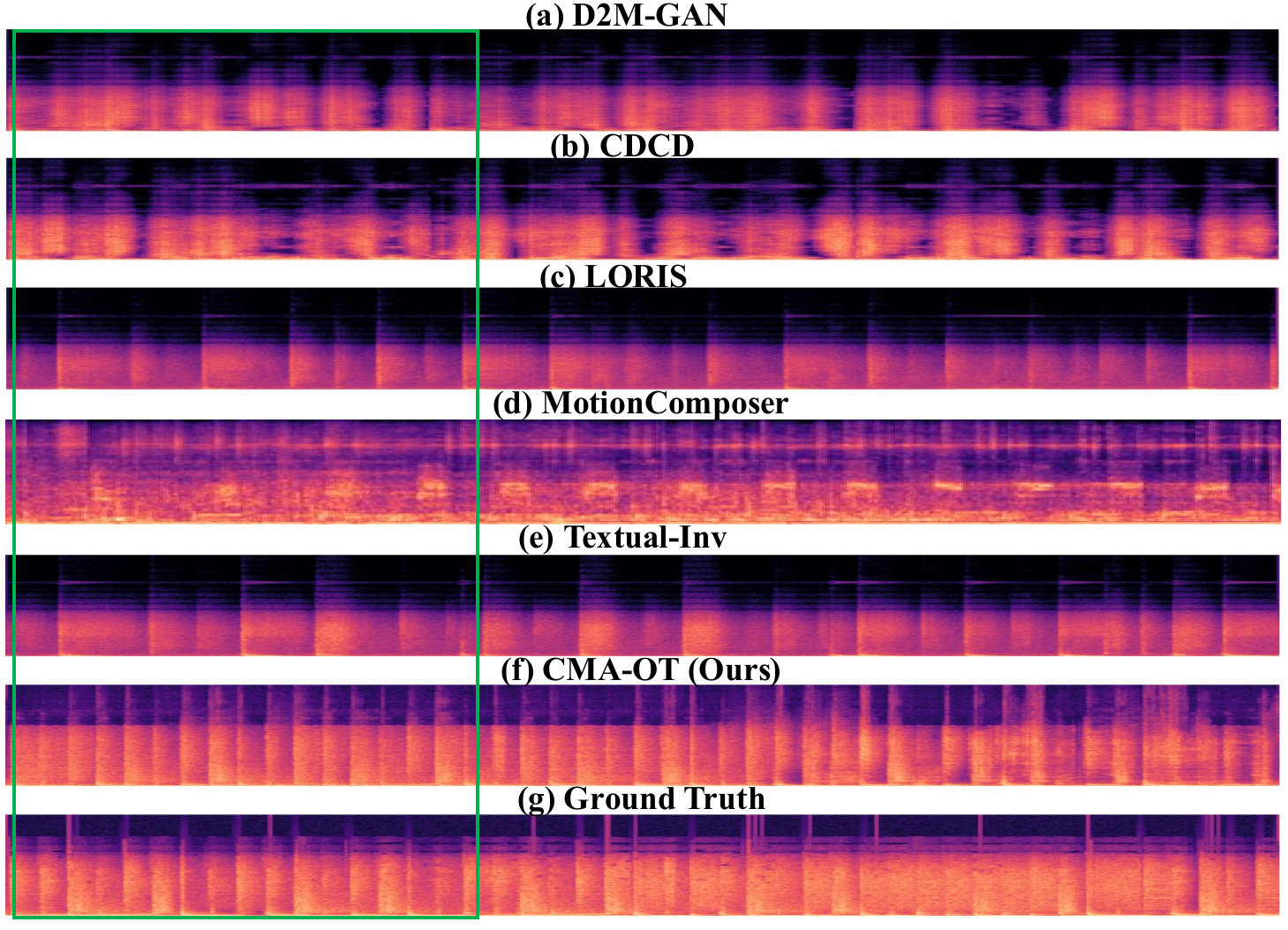}
    \caption{\textbf{Qualitative Comparison.} Our CMA-OT produces music with superior consistency closely matching the ground truth.}
    \label{fig:qualitative}
\end{figure}

\noindent\textbf{Qualitative Comparison.}
To further demonstrate the effectiveness of our CMA-OT, we conduct qualitative comparisons between CMA-OT and SOTA D2M methods. As shown in Figure~\ref{fig:qualitative}, CMA-OT generates music that exhibits superior rhythmic alignment and tonal coherence with the ground-truth music. In contrast, other competitive methods often produce temporally inconsistent or perceptually distorted musical outputs, which ultimately leads to poor synchronization between dance motions and the generated music.

\subsection{Ablation Study}

\noindent\textbf{Model Variants.}
To evaluate the contribution of each component in CMA-OT, we conduct ablation experiments on the AIST++ dataset, summarized in Table~\ref{tab:ablation}. We design eight variants to isolate the impact of external supervision, hierarchical alignment, curriculum learning, and scale-aware adaptation. The \textbf{Base} model serves as a baseline, employing a conditional flow matching framework without external representation alignment. \textbf{Base + CLAP-OT}~\cite{elizalde2023clap} and \textbf{Base + MERT-OT}~\cite{li2024mert} introduce external supervision by aligning DiT representations with CLAP or MERT embeddings via our scale-aware FGW. \textbf{Base + Top/Middle/Bottom-OT} applies single-scale FGW alignment at the top, middle, or bottom hierarchical levels of Jukebox. \textbf{Base + MA-OT} incorporates multi-scale FGW alignment without curriculum learning, activating all scales simultaneously. \textbf{Base + CMA} introduces curriculum-guided multi-scale alignment with cosine similarity loss, while \textbf{Base + CMA-OT*} combines curriculum-guided multi-scale alignment with FGW-OT but without scale-aware adaptive weighting. Finally, \textbf{Base + CMA-OT (Ours)} integrates curriculum-guided multi-scale alignment, scale-aware FGW-OT, and full hierarchical expert supervision.

\begin{figure}[t]
    \centering
    \includegraphics[width=0.99\linewidth]{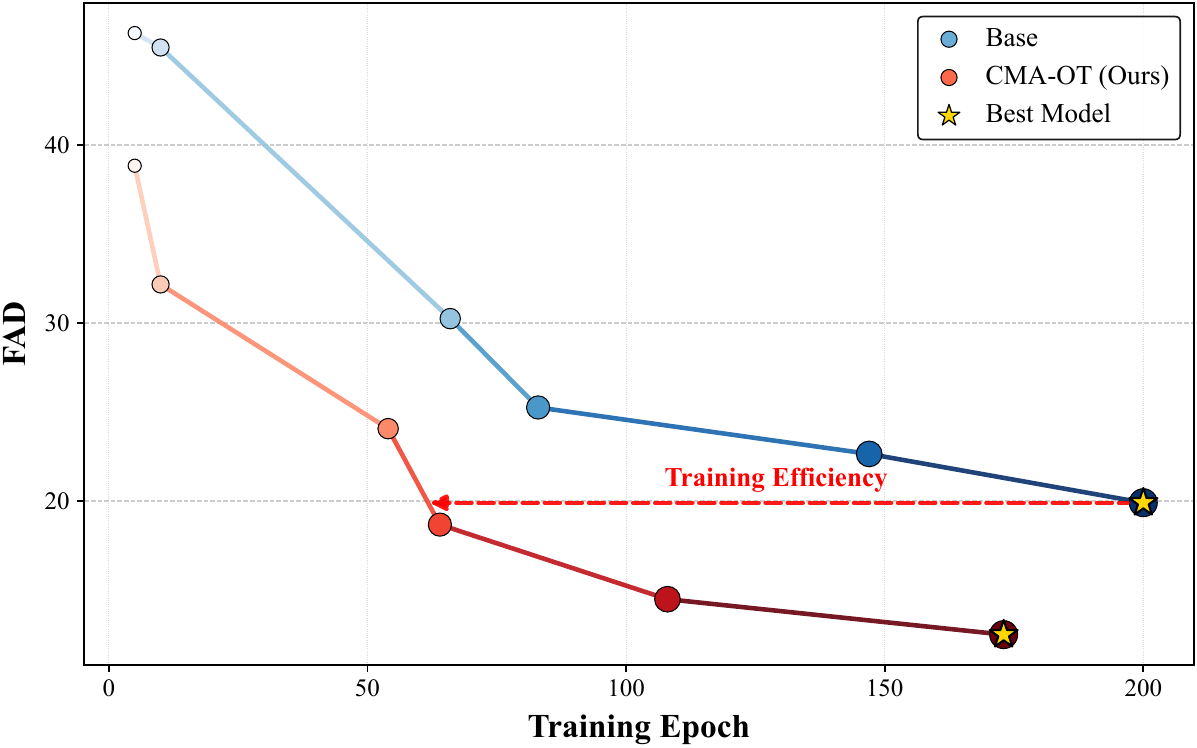}
   \caption{\textbf{Training Comparison.} Benefiting from multi-scale representation alignment, CMA-OT accelerates training and yields perceptually better music with significantly lower FAD than the baseline.}
    \label{fig:training}
\end{figure}

As illustrated in Table~\ref{tab:ablation}, representation supervision (Base + CLAP-OT and Base + MERT-OT) yields marginal improvements over the Base model, but global embeddings like CLAP and MERT provide limited guidance for fine-grained musical structure, which can impair rhythmic accuracy. In contrast, our single-scale expert alignment, particularly at the bottom level, captures detailed musical representations and achieves the largest rhythmic gains among single-scale variants. 

Multi-scale alignment without curriculum learning (Base + MA-OT) suffers from significant performance degradation, due to conflicting gradients across hierarchical scales, which destabilizes training. In contrast, curriculum-guided multi-scale alignment (Base + CMA) reduces FAD$_p$ by 40.81\% relative to the Base model, demonstrating that a coarse-to-fine curriculum effectively stabilizes training and enables reliable knowledge transfer across hierarchical levels.

Replacing cosine similarity with FGW (comparing Base + CMA and Base + CMA-OT*) improves all metrics, particularly fine-grained rhythmic alignment. This confirms that jointly modeling semantic similarity and temporal structure via FGW captures flexible dance-music correspondences more effectively than simple cosine similarity. Further incorporating scale-aware adaptive weighting (comparing Base + CMA-OT* and our CMA-OT) yields additional performance gains. This highlights that adaptive weighting optimally balances semantic and structural alignment across hierarchical levels and enhances multi-scale rhythm and style consistency. 
Our CMA-OT model achieves the best overall performance across all metrics, and it also accelerates training convergence.
Figure~\ref{fig:training} further shows that CMA-OT converges significantly faster than the baseline and other variants, indicating that our proposed paradigm enhances both training efficiency and final performance.

\noindent\textbf{Curriculum Learning Strategy.} 
To fully leverage hierarchical representations, we evaluate different curriculum learning strategies under the same CMA-OT architecture. We compare four schemes: fixed weights, where all scales are active simultaneously with static alignment weights; hard switch, a three-phase curriculum with abrupt sequential activation of each level; linear schedule, which gradually transitions between phases for smooth adaptation; and the proposed cosine schedule, which employs cosine-annealed weights for stable and progressive optimization. Details and a comparison visualization are provided in the supplementary material. 

As shown in Table~\ref{tab:curriculum_strategy}, the cosine schedule consistently outperforms other strategies, achieving the highest rhythmic and aesthetic scores while yielding the lowest FAD$_p$, demonstrating that smooth and continuous alignment emphasis during training effectively improves both optimization stability and generated music quality.

\begin{table}[t]
\centering
\caption{Comparison of curriculum learning strategy.}
\label{tab:curriculum_strategy}
\begin{tabular}{ccccccc}
\hline
 Method & BCS$\uparrow$ & F1$\uparrow$ & CE$\uparrow$ & FAD$_p$$\downarrow$  \\
\hline
Fixed Weights & 99.05 & 99.02 & 6.78 & 13.01  \\
Hard Switch & 99.07 & 99.04 & 6.82 & 12.79 \\
Linear Schedule & 99.12 & 99.09 & 6.86 & 12.57 \\
Cosine Schedule (Ours) & \textbf{99.14} & \textbf{99.12} & \textbf{6.93} & \textbf{12.50}  \\
\hline
\end{tabular}
\end{table}

\begin{table}[t]
\centering
\caption{Comparison of representation alignment methods.}
\label{tab:alignment_methods}
\begin{tabular}{ccccccc}
\hline
 Method & BCS$\uparrow$ & F1$\uparrow$ & CE$\uparrow$ & FAD$_p$$\downarrow$  \\
\hline
Linear+Cos & 98.83 & 98.79 & 6.68 & 13.69  \\
Time-Conv+Cos & 99.03 & 98.91 & 6.73 & 13.36 \\
 Standard OT & 99.05 & 98.97 & 6.75 & 13.05 \\
Scale-FGW (Ours) & \textbf{99.14} & \textbf{99.12} & \textbf{6.93} & \textbf{12.50}  \\
\hline
\end{tabular}
\end{table}

\noindent\textbf{Comparison of Temporal Alignment Strategies.} 
To validate the superiority of our scale-aware FGW alignment mechanism, we compare it with four representative alignment approaches, summarized in Table~\ref{tab:alignment_methods}. Baselines include: (1) \textbf{Linear+Cos}, simple linear projection with cosine similarity; (2) \textbf{Time-Conv+Cos}, time-convolutional projection with cosine similarity, as used in our Base + CMA variant; and (3) \textbf{Standard OT}, conventional optimal transport modeling only semantic similarity without structural constraints.  

As shown, cosine similarity-based methods provide only moderate improvements, as rigid point-wise matching cannot capture flexible temporal dynamics between dance and music. Standard OT improves performance by modeling distribution-level alignment but fails to preserve hierarchical structural consistency across scales. In contrast, our scale-aware FGW consistently achieves the best results across all metrics. These findings demonstrate that scale-aware FGW enables precise temporal matching between expert priors and generator states, effectively capturing multi-level dance-music correspondences and producing superior rhythmic consistency and high-fidelity music.



\subsection{User Study}

We conducted a user study for subjective evaluation, involving 40 volunteers who rated 30 test samples from the AIST++ dataset generated by six D2M methods. 

\begin{figure}[t]
    \centering
    \includegraphics[width=0.99\linewidth]{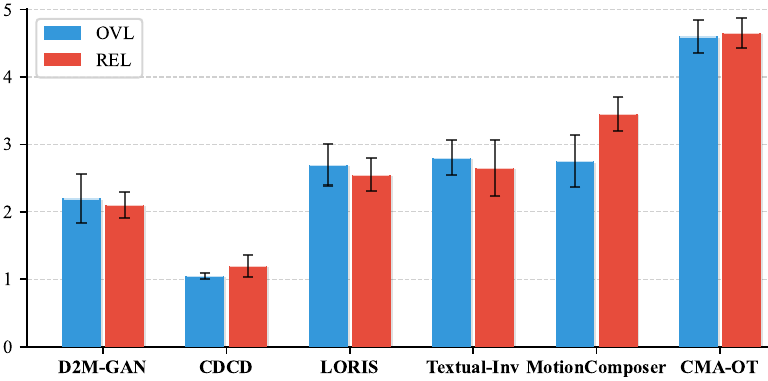}
\caption{\textbf{Subjective Evaluation.} User study results on rhythmic consistency and musical quality across  different D2M methods.}
    \label{fig:user_study}
\end{figure}

As shown in Figure~\ref{fig:user_study}, CMA-OT achieves the highest average scores in both metrics, consistently outperforming all baseline methods across the board. 
Notably, CMA-OT not only delivers significantly better dance-music synchronization, which aligns with our objective rhythm alignment metrics, but also generates far more coherent and aesthetically pleasing musical compositions, validating the effectiveness of hierarchical expert supervision in enhancing perceptual music quality.

\section{Conclusion}
\label{sec:conclusion}
In this work, we present \textbf{CMA-OT}, a curriculum-guided multi-scale alignment framework for D2M generation.
By introducing hierarchical expert supervision from a pre-trained multi-scale music expert, CMA-OT effectively bridges the fundamental semantic gap between sparse dance cues and dense musical semantics, directly addressing the core challenge of D2M generation.
The proposed curriculum-guided multi-scale learning strategy enables progressive knowledge transfer from high-level global musical structures to fine-grained temporal details, effectively alleviating gradient conflicts and ensuring stable cross-scale representation learning.
Meanwhile, the scale-aware FGW alignment mechanism models flexible correspondences between heterogeneous expert representations and generator latents, achieving precise, semantically consistent, and temporally coherent alignment across all hierarchical scales.
Extensive experiments on two datasets demonstrate that CMA-OT consistently outperforms existing SOTA methods in terms of rhythmic synchronization, perceptual quality, and overall music expressiveness.

\bibliographystyle{IEEEtran}
\bibliography{refer}

\appendix

\section{Multi-Scale Expert Feature Extraction}
\label{sec:feature_extraction}

To provide rich hierarchical supervision for dance-to-music (D2M) generation, we leverage a pre-trained generative music model, namely \textit{Jukebox 1B Lyrics}\footnote{https://huggingface.co/docs/transformers/model\_doc/jukebox}, using only its VQ-VAE encoder. The VQ-VAE encodes audio into three latent levels, \textbf{bottom}, \textbf{middle}, and \textbf{top}, capturing fine-grained rhythmic, melodic, and global stylistic information, respectively. Each level comprises 64-dimensional embeddings with a codebook of 2048 entries, and temporal downsampling is controlled via hop fractions of [0.125, 0.5, 0.5].  

All audio clips are first resampled to 44.1 kHz and converted to mono prior to encoding. The encoder outputs three discrete latent sequences, $[d_\text{bot}, d_\text{mid}, d_\text{top}]$, which are then mapped to continuous embeddings through the corresponding codebooks. The resulting features, denoted as 
\[
\mathcal{Z}_{\text{exp}} = \{\mathbf{z}_k\}_{k \in \{\text{top, mid, bot}\}},
\] 
have shapes $[T_\text{top}, 64], [T_\text{mid}, 64], [T_\text{bot}, 64]$, where $T_k$ depends on the input length and hop fraction.  
These continuous latent representations serve as hierarchical supervision signals for our D2M generator, providing coarse-to-fine guidance from global musical style to detailed temporal rhythmic patterns.

\section{Curriculum Learning Strategy}

To better illustrate the differences among the four curriculum learning strategies introduced in the main paper, we detail their mathematical formulations and corresponding weight evolutions across training epochs. All strategies control the relative weights of the three hierarchical alignment losses, denoted as $\lambda_{\text{top}}$, $\lambda_{\text{mid}}$, and $\lambda_{\text{bot}}$, corresponding to global, structural, and fine-grained rhythm levels, respectively.

\subsection{Fixed Weights}
In the fixed-weight baseline, the alignment weights remain constant throughout training:
\[
\lambda_{\text{top}} = 0.5, \quad \lambda_{\text{mid}} = 0.3, \quad \lambda_{\text{bot}} = 0.2.
\]
This setup activates all alignment levels simultaneously without any curriculum scheduling, serving as a conventional control configuration.

\subsection{Hard Switch}
The hard-switch curriculum divides the training into three distinct phases with abrupt transitions at epochs 30 and 100. Only one alignment level is emphasized in each phase:
\[
(\lambda_{\text{top}}, \lambda_{\text{mid}}, \lambda_{\text{bot}}) =
\begin{cases}
(1.0, 0.0, 0.0), & \text{if } \text{epoch} < 30, \\
(0.4, 0.6, 0.0), & \text{if } 30 \le \text{epoch} < 100, \\
(0.2, 0.3, 0.5), & \text{otherwise.}
\end{cases}
\]
While straightforward, such abrupt phase transitions often lead to optimization shocks and hinder stable convergence.

\subsection{Linear Schedule}
To alleviate the discontinuity of the hard switch, the linear schedule introduces smooth transitions between phases by linearly interpolating the weights:
\[
\lambda_t = (1 - \alpha_t)\lambda_{\text{start}} + \alpha_t \lambda_{\text{end}},
\quad \alpha_t = \frac{t - t_s}{t_e - t_s},
\]
where $t_s$ and $t_e$ denote the start and end epochs of each transition. This method enables gradual adaptation of alignment emphasis but still exhibits limited smoothness near transition boundaries.

\subsection{Cosine Schedule (Ours)}
Our cosine schedule further refines the transition process by applying a cosine annealing function:
\[
\lambda_t = \lambda_{\text{end}} + \tfrac{1}{2}(\lambda_{\text{start}} - \lambda_{\text{end}})\left[1 + \cos\left(\pi \tfrac{t - t_s}{t_e - t_s}\right)\right].
\]
This design ensures a smooth and continuous evolution of weights, enabling progressive and stable optimization across hierarchical scales. Specifically, the top-level weight $\lambda^{\text{top}}$ decays from 1.0 to 0.2, the mid-level weight $\lambda^{\text{mid}}$ rises from 0 to 0.6 and then decays to 0.3, and the bottom-level weight $\lambda^{\text{bot}}$ increases from 0 to 0.5 during later training.  
This progressive modulation enables the model to first establish coarse global alignment, then refine mid-level structures, and finally focus on fine-grained rhythm correspondence in a stable manner.

\subsection{Visualization}
\begin{figure*}[htbp]
    \centering
    \includegraphics[width=0.95\linewidth]{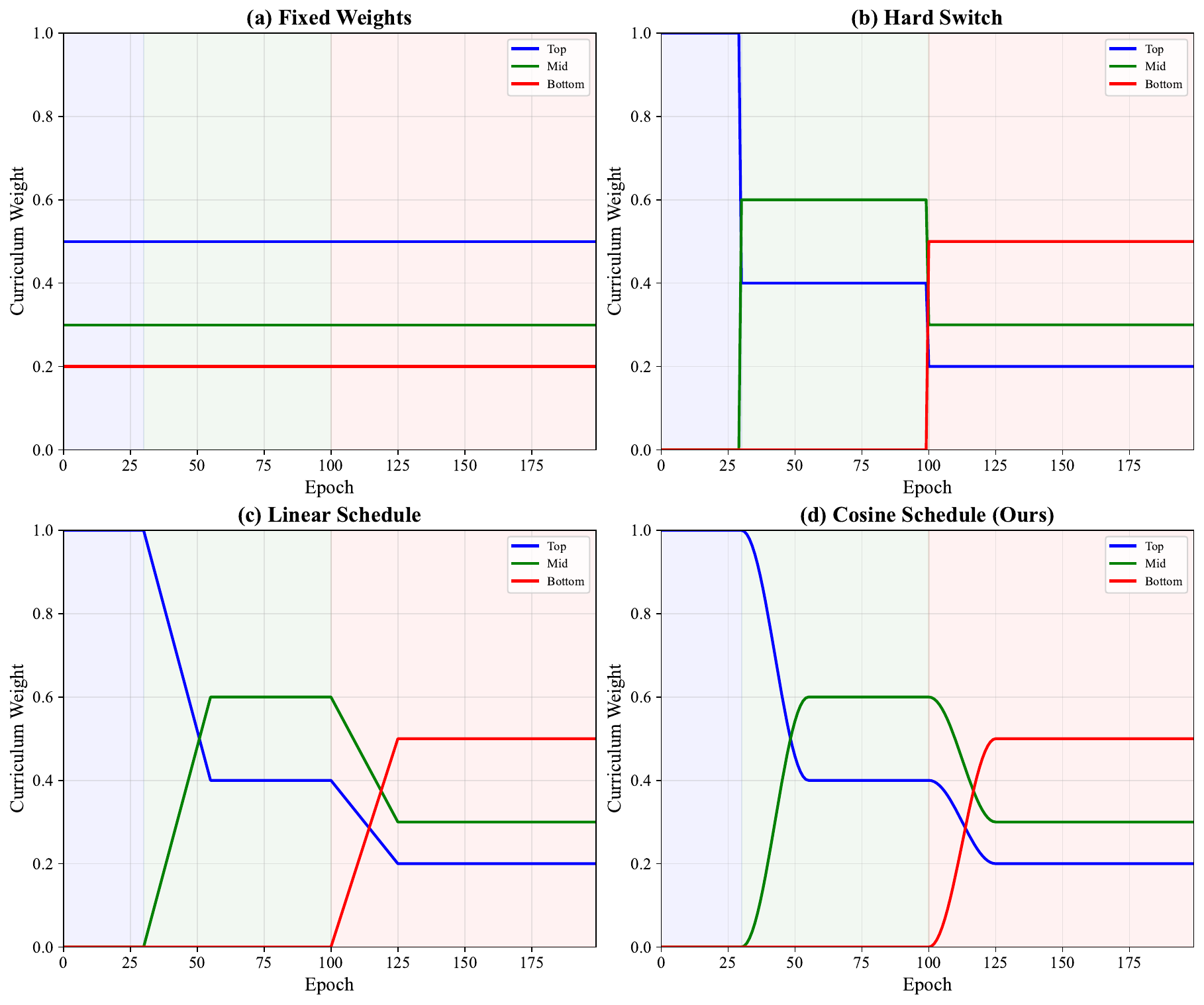}
    \caption{\textbf{Comparison of curriculum learning strategies.}
    The cosine-annealed schedule achieves smoother and more stable transitions across training epochs compared to fixed, hard-switch, and linear schemes, facilitating consistent optimization across hierarchical levels.}
    \label{fig:curri_supp}
\end{figure*}

Figure~\ref{fig:curri_supp} illustrates the evolution of the alignment weights for all four strategies. The cosine schedule provides the smoothest and most stable transitions, avoiding abrupt optimization shifts. In contrast, the hard-switch and linear schedules exhibit sharper transitions, while the fixed-weight baseline remains constant throughout training.

\section{Inference Details}
During inference, CMA-OT generates music conditioned on dance videos by extracting visual and rhythm features from the input video, which are then used to guide the conditional music generator. The generator, based on a Flow Matching Transformer (DiT), synthesizes the music latent, which is subsequently decoded into waveform using a pre-trained VAE. Notably, the representation guidance of external music expert employed during training is not used at inference; the generator operates  conditioned only on the extracted dance features, incurring no additional computational overhead compared to the base diffusion model.

\begin{figure*}[htbp]
    \centering
    \includegraphics[width=0.9\linewidth]{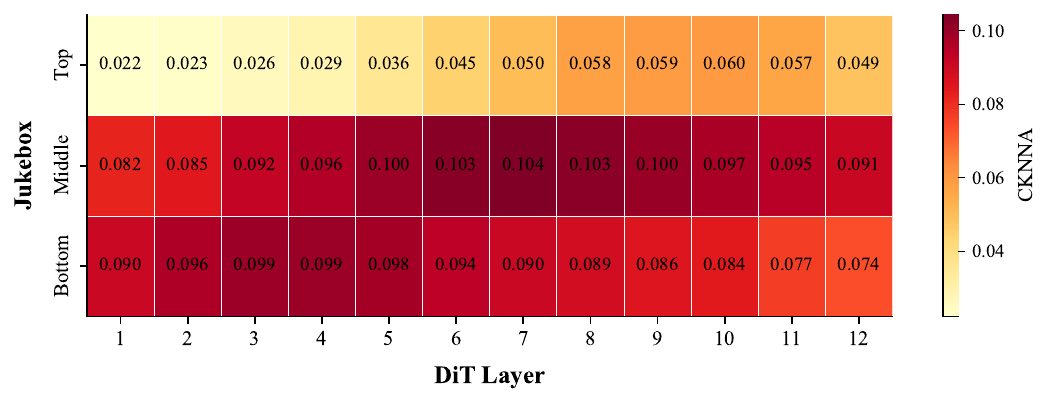}
    \caption{\textbf{Representation Similarity.} We provide CKNNA heatmap to show the representation similarity between DiT layers and the three Jukebox feature levels (bottom, middle, top).}
    \label{fig:heatmap}
\end{figure*}

\section{Layer Selection for Representation Alignment}
The effectiveness of our curriculum-guided multi-scale alignment relies on establishing meaningful correspondences between the hierarchical levels of the Jukebox expert and the internal layers of the DiT. To determine an optimal mapping, we conduct a representation similarity analysis between the DiT's internal states and Jukebox's hierarchical features.
\subsection{Analysis Method}
We perform the analysis on a subset of 1,000 clean music samples randomly drawn from the AIST++ training set. Each sample is forward-passed through the baseline model (without any alignment losses), and hidden representations from all 12 DiT layers are extracted at the final denoising step ($t=0$). We then compute the Centered Kernel Nearest-Neighbor Alignment (CKNNA)~\cite{pmlr-v235-huh24a} between these DiT representations and the three hierarchical levels of Jukebox features ($\mathbf{z}_{\text{top}}$, $\mathbf{z}_{\text{mid}}$, $\mathbf{z}_{\text{bot}}$) extracted from the same samples. We set $k=10$ following observations in \cite{pmlr-v235-huh24a} that smaller $k$ provides more reliable alignment.
\subsection{Analysis Results}
Figure~\ref{fig:heatmap} shows the resulting CKNNA similarity heatmap. A clear diagonal pattern is observed: bottom-level Jukebox features exhibit the highest similarity with shallow DiT layers, middle-level features align best with intermediate layers, and top-level features correspond to deeper layers. This pattern reflects a natural correspondence from fine-grained temporal details to global semantic structures, validating the hierarchical relationship between the two models' representations.
\subsection{Selected Layer}
Based on the analysis, we select DiT Layer 2 for bottom-level alignment, Layer 6 for middle-level alignment, and Layer 10 for top-level alignment. This principled mapping ensures semantically compatible representations are aligned at each scale, maximizing the effectiveness of hierarchical knowledge transfer from the Jukebox expert to the DiT generator.







\begin{figure}[htbp]
    \centering
    \includegraphics[width=0.99\linewidth]{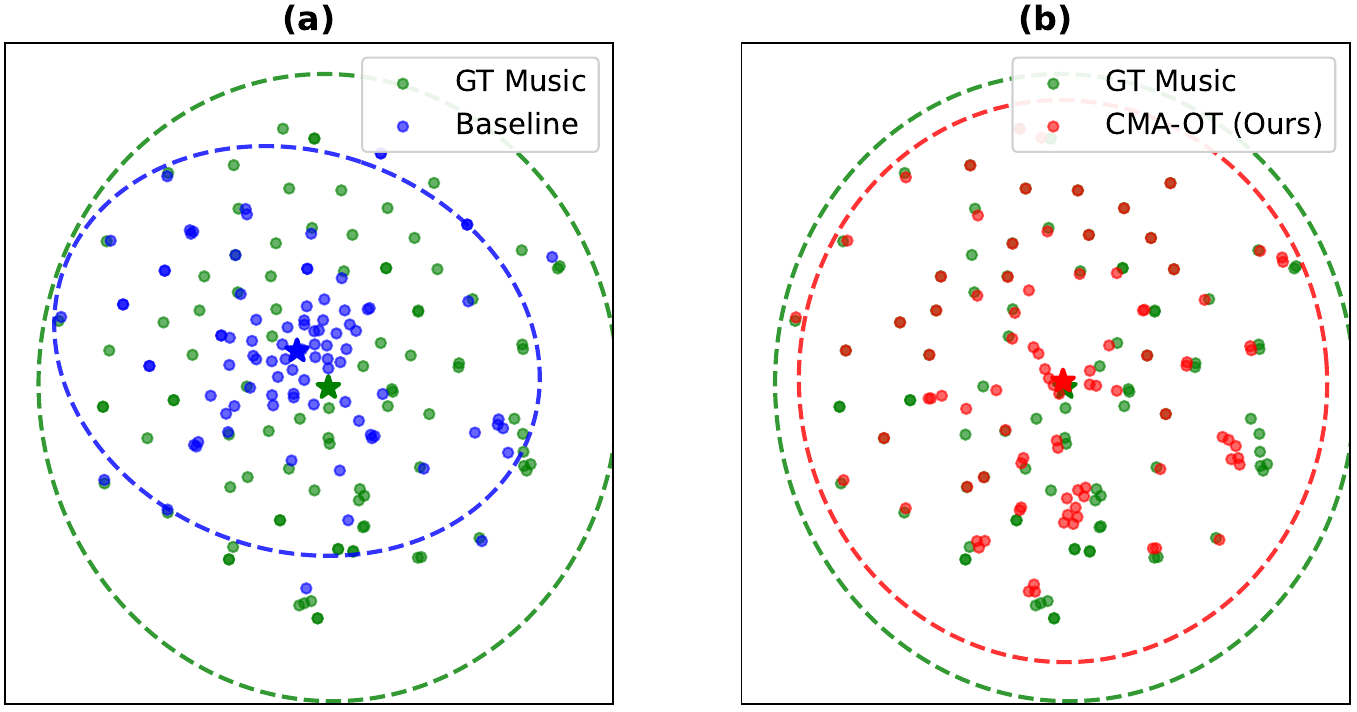}
    \caption{\textbf{Representation Distribution Comparison.} t-SNE visualization of music features from Baseline (a, blue), CMA-OT (b, red), and Ground Truth (GT) Music (green). Stars denote the mean of each distribution, and dashed ellipses indicate the 95\% confidence regions. }
    \label{fig:representation}
\end{figure}

\section{Representation Comparison.}
To further validate the effectiveness of CMA-OT, we analyze the distribution of generated music representations from the Baseline and our CMA-OT, visualized via t-SNE in Figure~\ref{fig:representation}.

In the Baseline setting (Figure~\ref{fig:representation}(a)), the generated feature distribution exhibits a clear spatial offset from the Ground Truth (GT) Music cluster, with limited overlap between their 95\% confidence ellipses and a significant gap between their distribution means. This reflects a severe semantic discrepancy between dance-conditioned generation and real music, a core limitation of the baseline framework.
In contrast, CMA-OT (Figure~\ref{fig:representation} (b)) produces feature representations that form a highly compact cluster, with its distribution mean nearly coinciding with that of GT Music and its confidence ellipse almost fully overlapping with the GT distribution.
These results demonstrate that CMA-OT effectively captures the semantic and rhythmic characteristics of real music, verifying that our hierarchical expert supervision successfully narrows the semantic gap between dance and music, significantly enhancing the representational quality of generated music.

\end{document}